\documentclass[runningheads]{llncs}

\usepackage{graphicx}

\usepackage{amsmath}
\usepackage{amssymb}
\usepackage{booktabs}

\usepackage[pagebackref,breaklinks,colorlinks]{hyperref}

\usepackage[capitalize]{cleveref}
\crefname{section}{Sec.}{Secs.}
\Crefname{section}{Section}{Sections}
\Crefname{table}{Table}{Tables}
\crefname{table}{Tab.}{Tabs.}

\begin{document}

\title{PE-former: Pose Estimation Transformer}
\author{Paschalis Panteleris\inst{1} \and
Antonis Argyros\inst{1,2}}
\authorrunning{P. Panteleris et al.}
\institute{Institute of Computer Science, FORTH, Heraklion, Crete, Greece \and
Computer Science Department, University of Crete, Greece
\email{\{padeler,argyros\}@ics.forth.gr}\\
\url{https://www.ics.forth.gr/hccv/}}
\maketitle  %

\begin{abstract}
Vision transformer architectures have been demonstrated to work very effectively for image classification tasks. Efforts to solve more challenging vision tasks with transformers rely on convolutional backbones for feature extraction. In this paper we investigate the use of a pure transformer architecture (i.e., one with no CNN backbone) for the problem of 2D body pose estimation. 
We evaluate two ViT architectures on the COCO dataset. We demonstrate that using an encoder-decoder transformer architecture yields state of the art results on this estimation problem.
\end{abstract}

\section{Introduction}
\label{sec:intro}
In recent years, transformers have been gaining ground versus traditional convolutional neural networks in a number of computer vision tasks. The work of Dosovitskiy et al~\cite{dosovitskiy2020image} demonstrated the use of the pure transformer model for image classification. This was followed by a number of recent papers demonstrating improved performance~\cite{touvron2021training}, 
reduced computational requirements~\cite{xiong2021nystr,wang2020linformer,el2021xcit,yuan2021tokens} 
or both. All of these pure transformer models are applied only to the task of image classification.
For more challenging vision tasks such as 2D human pose estimation, recent methods~\cite{li2021pose,yang2021transpose,mao2021tfpose} use a combination of a convolutional backbone followed by a transformer front-end to achieve similar performance to that of convolutional architectures.

In this work, we investigate the use of a pure transformer-based architecture for the task of 2D human pose estimation. More specifically, we focus on single instance, direct coordinate regression of the human body keypoints. We evaluate an encoder-decoder architecture derived from the work of Carion et al~\cite{carion2020end}. In our model we eliminate the need for a convolutional based backbone 
(a ResNet in the case of~\cite{carion2020end}) and we replace the encoder part of the transformer with a vision-based~\cite{dosovitskiy2020image} transformer encoder.
By foregoing with the need of a convolutional backbone, our model is realized by a much simpler architecture.

For the encoder part, we investigate the use of two vision transformer architectures:
\begin{itemize}
  \item Deit\cite{touvron2021training}: ViT architecture trained with distillation, which exhibits 
  baseline performance for vision transformers.
  \item Xcit\cite{el2021xcit}: Cross-covariance transformers. Xcit reduces the computational requirements by transposing the way ViT operates, i.e., instead of attending on tokens the transformer attends on the token-channels.
\end{itemize}

When comparing to other regression based methods~\cite{li2021pose,mao2021tfpose} 
which use a combination of transformers and convolutional backbones,
our Deit-based model performs on par with similar sized (in million parameters) models.
Moreover, our best model using an Xcit-based encoder 
achieves state of the art results on the COCO dataset.

We compare our transformer-only architecture with two ``baselines''. One that uses
a ResNet50 as the encoder section and another that uses a ViT (encoder+decoder) feature extractor.
The architecture and details of these baselines are discussed in Section~\ref{sec:method}.
We demonstrate experimentally that against these baselines, when everything else is kept the same, the transformer encoder improves performance.

Finally, we further improve the performance of our models using unsupervised pre-training, as proposed 
by Caron et al~\cite{caron2021emerging}. The training results of models that use supervised and unsupervised pre-training are presented in Section~\ref{sec:experiments}. 

In summary, the contributions of our work\footnote{Code is available on \href{https://github.com/padeler/PE-former}{https://github.com/padeler/PE-former}} are the following:
\begin{itemize}
  \item We propose a novel architecture for 2D human pose estimation that is using vision transformers 
  without the need for a CNN backbone for feature extraction.
  \item We demonstrate that our proposed architecture outperforms the SOTA on public 
  datasets by as much as $4.4\%$ in AP (compared to \cite{li2021pose}).
  \item We evaluate the performance of different vision transformer encoders for the 
  task of 2D human pose estimation.
  \item We demonstrate that the unsupervised pretraining of our transformer-based models, improves their performance by $0.5\%$ (for the Xcit model).
\end{itemize}

\begin{figure*}[t]
  \begin{center}
  \includegraphics[trim=60 0 60 0, clip, width=0.99\columnwidth]{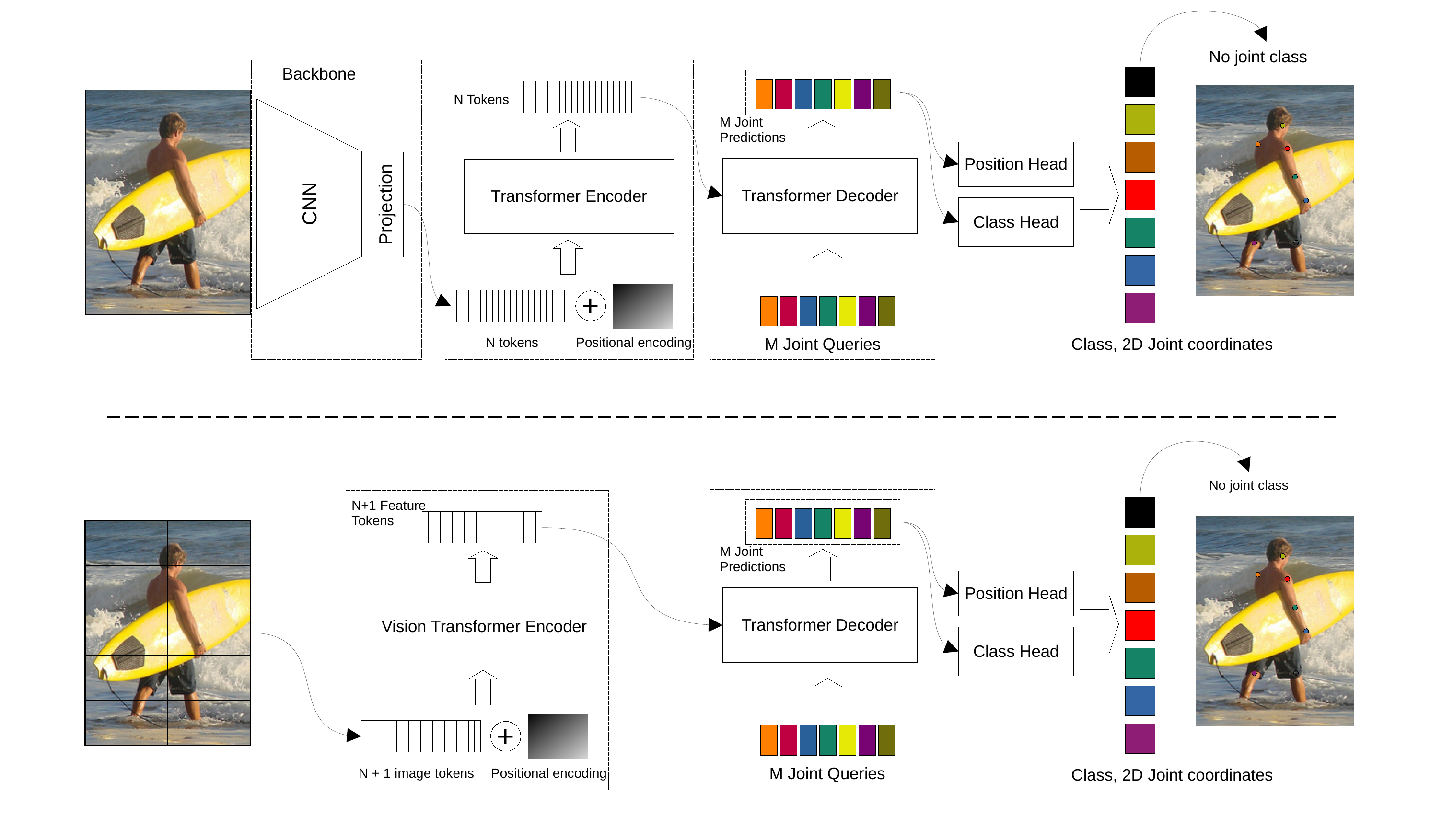}
  \caption{
  \label{fig:model}
  The DETR model adapted for 2D human pose estimation (top) and 
  our proposed model using Visual transformers (bottom).
  In our model, the cropped image of a person (bottom left), is split into $N$ tokens of $p \times p$ pixels
  (typically $p$ is equal to 8 or 16 pixels). The tokens are processed by the transformer encoder 
  (can be Deit or Xcit based). The output feature tokens ($N + 1$ for the class token) are used
  as the memory tokens in the DETR based transformer decoder. The decoder input is 
  $M$ joint queries. $M=100$ in our experiments. The Decoder outputs $M$ prediction tokens which are 
  processed by a classification and a regression head (FFNs). The output is 2D joint locations (range $[0,1]$) and class predictions.}
  \end{center}
\end{figure*}

\section{Related Work}
\label{sec:relatedwork}

\subsection{Vision Transformers}
The success of transformers in Natural Language Processing (NLP) has motivated a lot of researchers to adopt transformers for the solution of vision tasks. While attention-based mechanisms acting on image features such as the ones produced from a CNN, have been around for many years, only recently, 
Dosovitskiy et al~\cite{dosovitskiy2020image} demonstrated successfully with ViT the  
use of transformers for image feature extraction, replacing the use of CNNs for 
the task of image classification.
While the ViT approach is very promising, it still suffers from issues that arise from the 
quadratic complexity of the attention mechanism and translates to heavy computational and memory requirements.
Additionally, the vanilla ViT models and, especially, their larger variants, are very hard to train and require to be trained on huge annotated datasets.
Following ViT, many methods~\cite{xiong2021nystr,wang2020linformer,mehta2021mobilevit,yuan2021tokens,zhang2021aggregating}
appeared trying to solve or circumvent these issues while also 
maintaining SOTA performance~\cite{el2021xcit,touvron2021training}. 
In this work, we incorporate two very promising architectures, 
DEIT~\cite{touvron2021training} and Xcit~\cite{el2021xcit} in our models 
and evaluate their performance.

\subsection{Transformers for pose estimation}
Attention-based mechanisms have been applied to tackle demanding vision tasks  such as 2D human pose estimation. However, these rely on CNN backbones for feature extraction. 
The addition of an attention mechanism enables methods such as the one proposed by Li et al~\cite{li2021pose} to achieve state of the art results, improving on the best CNN based methods such as the HRNet by Sun et al~\cite{sun2019deep}.
However, Li et al Li et al~\cite{li2021pose} base their model and build on an HRNet CNN backbone. Similarly, methods such as 
TFPose by Mao et al~\cite{mao2021tfpose}, POET by Stoffl et al~\cite{stoffl2021end} and 
Transpose by Yang et al~\cite{yang2021transpose}, build on a robust CNN backbone and 
apply the attention mechanism on the extracted image features.

In this work, similarly to~\cite{li2021pose}, we use the transformer-based decoder module 
and the bipartite matching technique of Carion et al~\cite{carion2020end}. However, instead of relying on a CNN backbone or encoder for image features extraction, we introduce an architecture that directly uses the output features from a vision transformer.

\section{Method}
\label{sec:method}

As shown in Figure~\ref{fig:model} (bottom row), our architecture consists of two major components: A Visual Transformer encoder, and a Transformer decoder. The input image is initially converted into tokens following the ViT paradigm.
A position embedding~\cite{vaswani2017attention} is used to help retain the patch-location information.
The tokens and the position embedding are used as input to transformer encoder.
The transformed tokens are used as the memory~\cite{carion2020end} input of the transformer decoder.
The inputs of the decoder are $M$ learned queries.
For each query the network produces a joint prediction.
The output tokens from the transformer decoder are passed through two heads. The heads are feed forward neural networks (FFNs) following the architecture of DETR.
The first is a classification head used to predict the joint type (i.e., class) of each query.
The second is a regression head that predicts the normalized coordinates in the range $[0,1]$ 
of the joint in the input image.
Predictions that do not correspond to joints are mapped to a ``non object'' class.

In Section~\ref{sec:encoder} we discuss the details of our encoder module and 
we present the architecture of the various encoders that we examined. In Section~\ref{sec:decoder} we present the details of the decoder module which is 
derived from the DETR decoder. Finally, in Section~\ref{sec:training},
we discuss the training techniques and hyperparameters used for the experiments.

\subsection{Transformer Encoder}
\label{sec:encoder}
Dosovitskiy et al~\cite{dosovitskiy2020image} proposed the use of image patches which, in the case of ViT, have a size of $16 \times 16$. This approach of splitting the image into patches 
is adopted by all the vision transformer methods that appeared since ViT.
In general, an input image of dimensions $W \times H$ pixels, is split into $n \times m$ patches, each of which is a square block of $p \times p$ pixels.
These patches are flattened into vectors (tokens) and are subsequently passed through the transformer~\cite{vaswani2017attention} layers.  The self-attention mechanism operates on an input matrix $X \in R^{N \times d}$, where $N$ is the number of tokens. Each token has $d$ dimensions.
The input $X$ is linearly projected to queries, keys and values ($Q, K, V$).
Keys and values are used to compute an attention map 
$$A(K, Q) = Softmax(QK^{T}/ \sqrt{d}).$$ 
The results of the self-attention mechanism is the weighted sum 
of the values $V$ with the attention map: 
$$Attention(Q,K,V) = A(K,Q) V.$$
The computational complexity of this original self-attention mechanism scales quadratically with $N$,  due to pairwise interactions between all $N$ elements. 
The use of image patches in ViT instead of pixels makes the use of self-attention tractable, but 
the quadratic complexity still remains an issue.

In addition to the tokens created from the image patches, a vision transformer also has 
an extra (learned) token. This token, usually called the CLS token, is used to generate 
the final prediction when the transformer is trained for a classification task. 
The transformed tokens at the output of the encoder have the same number of channels as the input.
In our model, the output tokens of the encoder transformer including the 
CLS token are used  as the ``memory'' input of the decoder transformer.

\subsubsection{DeiT Encoder:}
\label{sec:deit}
Touvron et al~\cite{touvron2021training} proposed a training methodology for the ViT architecture 
that enables efficient training of the vision transformer without the use of huge datasets.
The proposed methodology enabled DeiT models to surpass the accuracy of the 
ViT baseline, while training just on the ImageNet dataset.
They further improved on that by proposing a distillation strategy that yielded accuracy 
similar to that of the best convolutional models.
Our DeiT encoder uses the \emph{DeiT-S} model architecture. For our experiments, the encoders are initialized using either the weights provided by the DeiT authors (DeiT-S, trained with distillation at $224 \times 224$) of the weights provided by the Dino authors.
For our experiments we use two versions of the DeiT-S model, both trained with 
input sizes of $224 \times 224$ pixels:
\begin{enumerate}
\item
the DeiT-S variant trained with distillation and
\item
the DeiT-S trained using the unsupervised methodology of Dino.
\end{enumerate}

\subsubsection{Xcit Encoder:}
El-Nouby et al~\cite{el2021xcit} recently proposed the Xcit transformer. 
In this approach they transpose the encoder architecture. 
Attention happens between the channels of the tokens rather than between
the tokens themselves. The proposed attention mechanism is complemented by 
blocks that enable ``local patch interations'', enabling the exchange of information 
between neighboring tokens on the image plane.

The Xcit transformer scales linearly with the size of patches since the 
memory and computation expensive operation of attention is applied to the channels 
which are fixed in size. This characteristic of Xcit enables training with larger input image sizes and more patches. In our experiments we use the \emph{Xcit-Small-12} variant of the model
which corresponds roughly to the \emph{DeiT-S} variant of the DeiT architecture.
Similar to the DeiT (Section~\ref{sec:deit}) encoder we use pre-trained weights for this 
encoder. The weights are obtained with 
\begin{enumerate}
\item
supervised training on the ImageNet and 
\item
with the unsupervised methodology of Dino.
\end{enumerate}

\subsubsection{Resnet Encoder:}
In order to verify the positive impact of our proposed attention-based encoder
we implemented a third type of encoder, this time using Resnet50~\cite{he2016deep} 
as the architecture. The 50-layer variant was selected because it has a similar number 
of parameters as the examined transformer-based encoders.

The Resnet encoder replaces the ``Visual Transformer Encoder'' block 
shown in Figure~\ref{fig:model}. Following the DETR implementation we use a $1 \times 1$ convolution to project the output features of the Resnet to the number of channels expected by the decoder. Subsequently we unroll each feature channel to a vector. These vectors are
passed as the memory tokens to the decoder.
The Resnet50 encoder is used to experimentally validate the positive effect of 
using vision transformers (DEIT or Xcit) versus a conventional convolutional encoder. 

\subsubsection{VAB (ViT as Backbone):}
Related work that tackles pose estimation with transformers typically uses a convolutional backbone 
to extract image features. This model architecture is shown on the top row of Figure~\ref{fig:model}. The features created by the CNN backbone are flattened and used as input tokens to a transformer front-end. In this model the transformer consists of an encoder and a decoder block.

For our experiments we replace the CNN backbone for a visual transformer.
The vision transformer outputs feature tokens which are subsequently processed by the 
(unchanged) transformer encoder-decoder block. We call this modification VAB for ``ViT as Backbone''. 
The VAB approach is closer in spirit to contemporary methods for pose estimation 
replacing only the CNN backbone with ViT. However it is larger in number of parameters 
and requires more computational resources than its CNN-based counterparts.
In our experiments we use the DeiT-S as the visual transformer backbone.

\subsection{Transformer Decoder}
\label{sec:decoder}
We adapt the decoder block of DETR and use it in our model. A number of learned queries are used to predict joint locations. Predictions that do not correspond to a joint are mapped to the ``non-object'' class. Following the work of Li et al~\cite{li2021pose} we use $M=100$ queries for all our models. Adding more queries does not improve the result, while having queries equal to the number of  expected joints (i.e., $17$ for the COCO dataset) gives slightly worse results.

Bipartite graph matching is used at training time  to map the regressed joints to the 
ground truth annotations. In contrast to the DETR decoder, our decoder is 
configured to regress to 2D keypoints rather than object bounding boxes. 
This translates to regressing $2$ scalar values ($x,y$) in the range of $[0,1]$ for each joint. In all our experiments we used a decoder with $6$ layers which achieves a good performance balance. Adding more layers to the decoder gives slightly improved results at a cost of more parameters and higher computational and memory requirements.

\subsection{Training}
\label{sec:training}
We use AdamW to train our models with a weight decay of $1e-4$. For all our experiments we use pretrained encoders while the decoder weights are randomly initialized. The pretrained weights used are noted in each experiment. The learning rate for the encoder is set to $1e-5$ and for the decoder to $1e-4$. The learning rate drops by a factor of $10$ after the first $50$ epochs and 
the models are trained for a total of $80$ epochs. For all our models we use a 
batch size of $42$.

For data augmentation we follow the approach of Xiao et al~\cite{xiao2018simple}. According to this we apply a random scaling in the range $[0.7, 1.3]$, a random rotation in the range ($[-40^\circ, 40^\circ]$) and a random flip to the image crop. We  found experimentally that the jitter, blur and solarization used by~\cite{carion2020end} for the object detection task were not helpful in our 2D human pose estimation problem, so these operations were not used.

\section{Experiments}
\label{sec:experiments}
We compare against two other regression based methods~\cite{li2021pose,mao2021tfpose} for 
single instance 2D human pose estimation. Both of these methods use a convolutional backbone 
and an attention mechanism derived from the DETR decoder to regress to the 
2D keypoints of a human in the input image crop.

\subsection{Evaluation methodology}
We train and evaluate our models on the COCO val dataset. We use the detection results from a person detector with AP 50.2 on the COCO Val2017 set. Following standard practice~\cite{xiao2018simple,li2021pose} we use flip test and average the results from the original and flipped images. Unless otherwise noted, the methods we compare against use the same evaluation techniques.

\subsection{Comparison with SOTA}
As shown in Table~\ref{tab:cmp}, our architecture performs on par or surpasses methods that use CNN backbones for the same task. Using the DEIT encoder requires more memory and CPU resources although the number of parameters is relatively low. Due to the increased requirements, we limit our comparison tests of DEIT to $192 \times 256$ resolution for patch sizes of $8 \times 8$ pixels. 

Our DEIT variant scores just $0.4\%$ lower than TFPose
\footnote{We contacted the authors of TFPose for additional information to use in our comparison, such as number of parameters and AR scores but got no response}
for input $192 \times 256$. 
However, it outperforms PRTR @ $288 \times 384$, despite the higher input resolution and the 
greater number of parameters ($5.1M$) of the later.

Due to its reduced computational and memory requirements, our Xcit variant  can be trained 
to higher resolutions and, thus, larger number of input tokens. We train networks at $192 \times 256$ and $288 \times 384$ input resolutions for the  $16 \times 16$ and $8 \times 8$ patch sizes. Our Xcit based models outperform both the TFPose and PRTR networks at their corresponding resolutions, while having the same number of parameters as PRTR. 

Overall, patch size of $8 \times 8$ pixels and resolutions of $288 \times 384$ yield the best performance. In fact, our Xcit-based network (Xcit-dino-p8) at $288 \times 384$ outperforms the PRTR trained at the same resolution by $4.4\%$ in AP and $3.4\%$ in AR and even outperforms PRTR at resolution $384 \times 512$ by $1.6\%$ in AP. However, experiments of the Xcit variant at even higher resolutions ($384 \times 512$) did not show any significant improvements.

\begin{table}[t]
  \centering
  \begin{tabular}{|l||r|r|r|r|}
   \hline 
   {\bf Method} & {\bf Input size} & {\bf \#Parameters} & {\bf AP} & {\bf AR} \\
   \hline\hline
   TFPose       & 192x256 & --    & 71.0 & -- \\
   \hline
   TFPose       & 288x384 & --    & 72.4 & -- \\
   \hline
   PRTR         & 288x384 & 41.5M & 68.2 & 76 \\ 
   \hline
   PRTR         & 384x512 & 41.5M & 71.0 & 78 \\ 
   \hline
   OURS-Deit-dino-p8  & 192x256 & 36.4M & 70.6 & 78.1 \\
   \hline
   OURS-Xcit-p16      & 288x384 & 40.6M & 70.2 & 77.4 \\
   \hline
   OURS-Xcit-dino-p16 & 288x384 & 40.6M & 70.7 & 77.9 \\ 
   \hline
   OURS-Xcit-dino-p8  & 192x256 & 40.5M & 71.6 & 78.7 \\ 
   \hline
   OURS-Xcit-dino-p8  & 288x384 & 40.5M & 72.6 & 79.4 \\ 
   
   \hline
  \end{tabular}
  \caption{A comparison of transformer based methods for 2D body pose estimation with direct regression. Both TFPose~\cite{mao2021tfpose} and PRTR~\cite{li2021pose} models use a Resnet50 for backbone. The dataset is COCO2017-Val. Flip test is used on all methods. The decoder depth for all models is $6$. Note: the code for TFPose~\cite{mao2021tfpose} is not available and the model is not provided by the authors, so the number of parameters is unknown. In our work the ``-Dino'' suffix denotes the use of unsupervised pretraining, Xcit refers to Xcit-small-12, and Deit refers to Deit-small.}
  \label{tab:cmp}
\end{table}

\begin{table}[t]
  \centering
  \begin{tabular}{|l||r|r|r|}
   \hline 
   {\bf Method} &  {\bf \#Parameters}  & {\bf AP} & {\bf AR} \\
   \hline\hline
   Resnet50-PE-former   & 39.0M & 63.4 & 72.2 \\ 
   \hline
   VAB             & 47.4M & 63.2 & 72.6 \\ 
   \hline
   Deit             & 36.4M & 62.2 & 71.6 \\ 
   \hline
   Xcit            & 40.6M & 66.2 & 74.6 \\ 
   \hline
  \end{tabular}
  \caption{Encoder Comparison. Deit vs Xcit vs resnet50 vs VAB. 
  For all experiments the patch size is set to $16 \times 16$ pixels. Input resolution is $192 \times 256$.
  All Networks are trained and evaluated on the COCO val dataset. Deit performs worse while also having the smallest number of parameters. Xcit is the best performing overall, however it also has $10\%$ more parameters than Deit.}
  \label{tab:enc_cmp}
\end{table}

\begin{table}[t]
  \centering
  \begin{tabular}{|l||c|c|c|}
   \hline 
   {\bf Method} &  {\bf \# Parameters} & {\bf AP} & {\bf AR} \\
   \hline\hline
   Deit               & 36.4M & 62.2 & 71.6 \\ 
   \hline
   Dino Deit          & 36.4M & 66.7 & 75.0 \\ 
   \hline
   Xcit               & 40.6M & 66.2 & 74.6 \\ 
   \hline
   Dino Xcit          & 40.6M & 68.0 & 76.1 \\ 
   \hline\hline
   Resnet50-PE-former      & 39.0M & 63.4 & 72.2 \\ 
   \hline
   Dino Resnet50-PE-former & 39.0M & 61.0 & 70.1 \\ 
   \hline
  \end{tabular}
  \caption{Comparison of Unsupervised Dino weight init. 
  Patch size is set to $16$ and input size is $192 \times 256$. Dino networks are 
  identical but initialized with the weights of networks trained  with the dino unsupervised methodology.   Networks are trained and evaluated on the COCO val dataset. Interestingly, the Resnet50 variant does not show the same improvements as the attention-based encoders with unsupervised pre-training.}

  \label{tab:unsupervised}
\end{table}

\subsection{Transformer vs CNN vs VAB Encoders}
\label{sec:enc_cmp}
We evaluate the performance of a transformer based encoder (Deit, Xcit) versus a 
CNN (Resnet50) based encoder and a VAB model. The networks are trained on COCO 
and evaluated on the COCO2017-val. 
For all the encoders, the imagenet pretrained weights are used. The results are shown in Table~\ref{tab:enc_cmp}. The networks are trained with an input resolution of $192 \times 256$ and patch size of $16 \times 16$ pixels.
Apart from replacing the encoders, the rest of the model (decoder) and training hyperparameters are kept the same.

As expected, VAB is the largest network with the highest memory and computational requirements. 
However, it performs on par with the Resnet50 variant. The Resnet variant despite having almost the same number of parameters as Xcit, requires less memory and computational resources during training and inference. Still, it under-performs compared to Xcit by $2.8\%$ in AP and $2.4\%$ in AR.
The Deit variant performs a bit worse ($1.2\%$ in AP) than Resnet but it is also smaller by $2.6M$ parameters. 

Xcit exhibits the best overall performance with a lead of $2.8\%$ in AP and $2.4\%$ in AR over the Resnet50 variant.

\subsection{Unsupervised pre-training}
\label{sec:exp_unsupervised}
All the models presented in Section~\ref{sec:method} use pre-trained 
weights for the encoder part, while the decoder's weights are randomly initialized.
We use the weights as they are provided by the authors of each architecture 
Deit\footnote{\url{https://github.com/facebookresearch/deit}},
Xcit\footnote{\url{https://github.com/facebookresearch/xcit}}). 
Additionally, we evaluate the use of weights\footnote{\url{https://github.com/facebookresearch/dino}}
created using unsupervised learning~\cite{caron2021emerging}.
More specifically, we evaluate using the Deit, Xcit and Resnet50 encoders.

Table~\ref{tab:unsupervised} presents the obtained results.
For the purposes of our evaluation we train two Deit (deit-small) and two Xcit (xcit-small-12) variants. For all variants, input resolution of $192 \times 256$ and patch size is set to $16 \times 16$. All hyperparameters are kept the same. We initialize the encoders of the Dino 
variants using weights acquired with unsupervised learning on Imagenet. In contrast, the normal variants start with encoders initialized with weights acquired with supervised learning on Imagenet. For both Deit and Xcit we observe significant improvement on the performance. 
However, the amount of improvement drops as the overall performance of the network gets higher (see the entries of Table~\ref{tab:cmp} for \emph{OURS-Xcit-p16} and \emph{OURS-Xcit-dino-p16}).

As the authors of~\cite{caron2021emerging} hypothesise, we believe that this improvement 
stems from having the dino-trained model learn a different embedding than the supervised variant. 
The model is focusing only on the salient parts of the image (i.e., the person) and is not miss-guided by  annotation biases and errors that affect the supervised approach. This hypotheses is further supported by the recent work of He et al~\cite{he2021masked} on unsupervised learning using masked autoencoders.

Interestingly, our CNN based encoder (Resnet50) does not benefit from the unsupervised pretraining. The resnet50-PE-former-dino pretrained with the DINO weights, yields worse results than the supervised
variant (resnet50-PE-former). This result hints to a fundamentally different way the transformer-based methods learn about salient image parts:
When training a CNN, interactions are local on each layer (i.e., depend on the size of the convolutional kernel). 
On the other hand, vision transformers such as Xcit and Deit, allow for long-range interactions. Long range interactions enable the exchange of information between areas in the image that are far apart. For DINO the implicit task during training is to identify salient parts in an image. As a consequence long range interactions help the network reason about larger objects (i.e, a person).

For models with the transformer based encoders (deit, xcit), it is possible that further fine tuning such as pre-training on the same resolution as the final network or pre-training on a person dataset (i.e COCO-person) instead of imagenet, could further boost the final gains. 
However these tasks are beyond the scope of this work and are left for future work.

\section{Conclusions}
We presented a novel encoder-decoder architecture using only transformers that achieves SOTA results for the task for single instance 2D human pose estimation. We evaluated two very promising ViT variants as our encoders: Xcit and Deit. We verified the positive impact of our transformer based encoder-decoder by comparing with modified versions of our model with a CNN-encoder and a VAB variant.
Out model using the Xcit based encoder, performs best both in accuracy and resource requirements, outperforming contemporary methods by as mach as $4.4\%$ in AP on the COCO-val dataset.
Our Deit based encoder variant is on par with methods using CNN based backbones for patch sizes of $8 \times 8$ pixels.
Furthermore, we demonstrate that using unsupervised pretraining can improve performance, especially for larger patch sizes (i.e., $16 \times 16$). 

Attention-based models look promising for a range of vision tasks.
It is conceivable that, with further improvements on the architectures or 
the training methodologies, these models will dethrone the older CNN-based 
architectures both in accuracy and resource requirements.

\bibliographystyle{splncs04}
\bibliography{report}

\end{document}